\documentclass{article}
\usepackage[T1]{fontenc}
\usepackage[utf8]{inputenc}
\newif\ifarxiv
\arxivtrue   

\usepackage{ismir} 

\ifarxiv
  \renewcommand{\permission}{}
\fi

\usepackage{amsmath,cite,url}
\usepackage{graphicx}
\usepackage{color}
\usepackage{multirow}
\usepackage{makecell}
\usepackage{booktabs}
\usepackage{svg}
\usepackage{subcaption}
\usepackage{enumitem}
\usepackage{caption}
\usepackage{subcaption}
\title{KARMA-MV: A Benchmark for Causal Question Answering on Music Videos }

\ifarxiv
  \usepackage{authblk}
\fi
\ifarxiv
 \author{
  \parbox{\textwidth}{\centering
    Archishman Ghosh \quad Abhinaba Roy \quad Dorien Herremans \\
    AMAAI Lab, Singapore University of Technology and Design \\
    \small \texttt{archishman\_ghosh@mymail.sutd.edu.sg}, \texttt{\{abhinaba\_roy,dorien\_herremans\}@sutd.edu.sg}
  }
}
\else
    \oneauthor
     {Archishman Ghosh, Abhinaba Roy, Dorien Herremans}
     {AMAAI Lab\\ Singapore University of Technology and Design\\ \small \texttt{archishman\_ghosh@mymail.sutd.edu.sg},\\ \small \texttt{\{abhinaba\_roy,dorien\_herremans\}@sutd.edu.sg}}
\fi

\def\authorname{A. Ghosh, A. Roy, and D. Herremans}

\renewcommand{\permission}{}  
\begin{document}

\maketitle

\begin{abstract}
While significant progress has been made in Video Question Answering and cross-modal understanding, causal reasoning about how visual dynamics drive musical structure in music videos remains under-explored.We introduce KARMA-MV, a large-scale multiple-choice QA dataset derived from 2,682 YouTube music videos, designed to test models' ability to integrate temporal audio-visual cues and reason about visual-to-musical influence across reasoning, prediction, and counterfactual questions. Unlike traditional datasets requiring manual annotation, KARMA-MV leverages LLM reasoning for scalable generation and validation, yielding 37,737 MCQs.

We propose a causal knowledge graph (CKG) approach that augments vision-language models (VLMs) with structured retrieval of cross-modal dependencies. Experiments on state-of-the-art VLMs and LLMs show consistent gains from CKG grounding—especially for smaller models—establishing the value of explicit causal structure for music-video reasoning. KARMA-MV provides a new benchmark for advancing causal audio-visual understanding beyond correlation. 
\end{abstract}

\section{Introduction}\label{sec:introduction}
Audio and video are two dominant perceptual channels through which human experience the world, and understanding how these modalities co-evolve is fundamental to a broad class of multimedia reasoning problems. When watching a music video by famous rock bands or a choreographed dance sequence from a film, it is natural to observe tight coordination between what unfolds visually and how the music responds. Scene transitions align with chord progression, intensity  of action correlates with a change in loudness, etc.  Investigating the precise nature of these causal dependencies, specifically how changes in visual features drive changes in musical features, is both scientifically interesting and practically important for building systems that truly understand audio-visual content. To that end, we present a structural framework for characterizing the causal relationship between video and music. Since we are primarily concerned with music videos, we focus on the musical features that change in response to shifts in video features. Throughout this work, ``causal'' refers to the directional visual-to-audio framing, i.e., how visual changes systematically  precede and predict musical changes, rather than interventional causality in the formal sense.In this paper, we introduce \textbf{KARMA-MV}, a Causal Music-Video Question Answering MCQ dataset based on the scenes from music videos, targeting the causal effect of visual changes on musical and audio features. 

The main focus of KARMA-MV is on causal influence of video on music, \textit{i.e.}, how musical properties change when video features change. To construct the dataset, we extract audio and visual features independently and feed them into the state-of-the-art Large Language Model Qwen2.5-7B-Instruct~\cite{qwen2.5,qwen2_tech} to generate semantically grounded question-answer pairs. The questions are designed to be interpretable by domain experts and to require genuine cross-modal reasoning rather than unimodal shortcuts.

\begin{figure*}[htpb]
    \centering
    \includegraphics[width=0.75\textwidth]{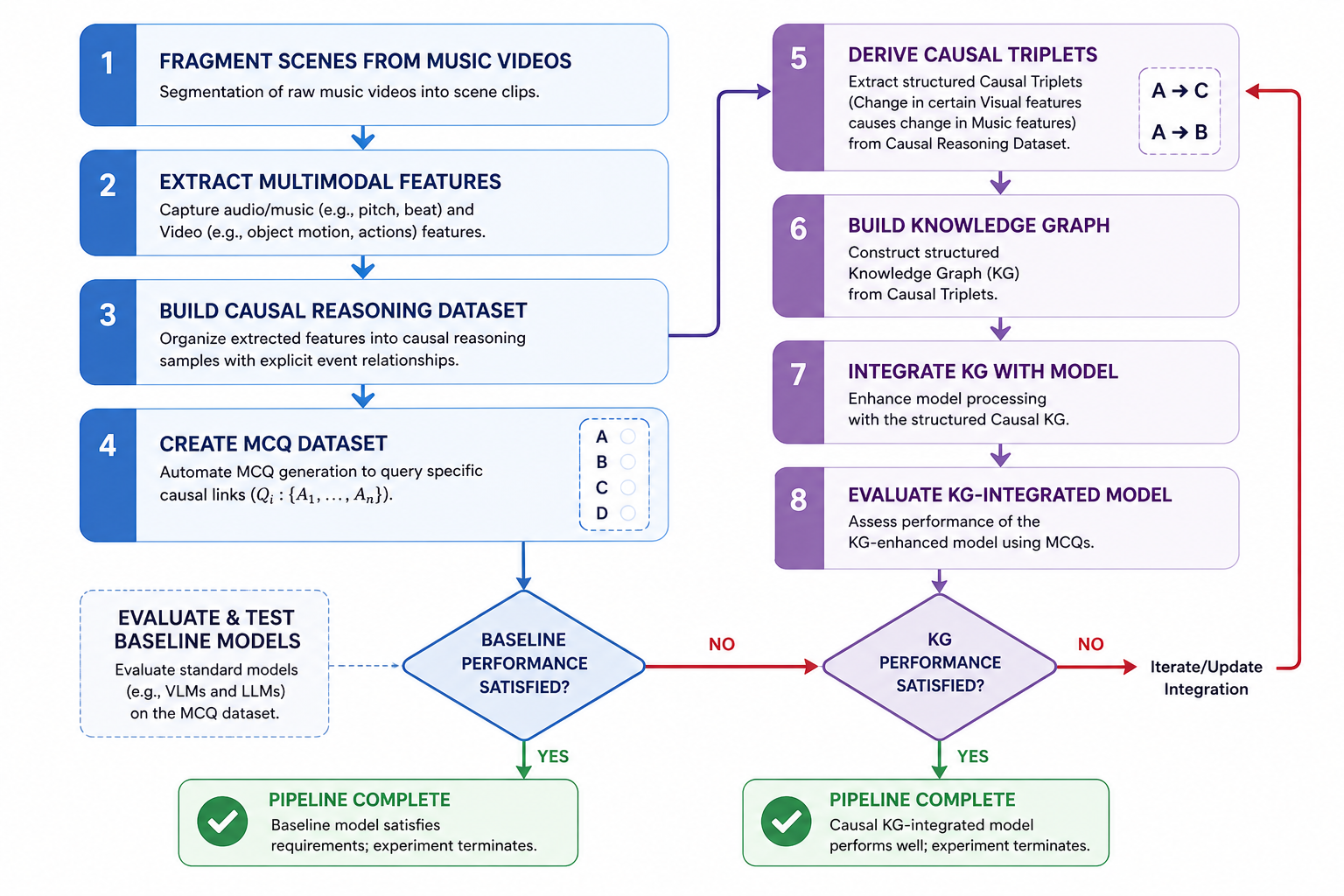}
\vspace{-6pt}
    \caption{\footnotesize Overview of the proposed KARMA-MV approach. }
    \label{fig:overview}
\vspace{-12pt}
\end{figure*}

Following Li et al.~\cite{li2022representation}, KARMA-MV evaluates whether models can understand what is there in the video (description), explain the intentions of actions or procedures to certain targets (explanation), predict what may happen in the future (prediction), and imagine the scenarios in different conditions (counterfactual). However, unlike Li et al.~\cite{li2022representation}, we focus solely on the visual content; our benchmark is explicitly grounded in how visual changes causally influence musical properties. We validate our work against two Vision Language Models capable of jointly processing text, video and audio. 

\noindent\textbf{Contributions.} The primary contributions of this work are as follows:
\begin{enumerate} [nosep, leftmargin=*] 
    \item We introduce a large-scale, automatically generated MCQ dataset for causal reasoning in music videos, covering description, explanation, prediction, and counterfactual question types across diverse audio-visual scenes.
    \item We propose an LLM-driven pipeline for dataset construction that eliminates the need for manual annotation while preserving semantic complexity, enabling straightforward extension to new video sources.
    \item We design an architecture that integrates a Causal Knowledge Graph with a Vision Language Model, enabling structured cross-modal reasoning over audio-visual dependencies.
    \item We provide a comprehensive evaluation of state-of-the-art VLM baselines on KARMA-MV and demonstrate that explicit causal modeling via knowledge graphs yields consistent improvements, establishing a strong baseline for future work on causal music-video understanding.

\end{enumerate}

In the remainder of this paper, we describe how a Causal Knowledge Graph based VLM architecture can be constructed for Music-Video Question Answering task. We examine in detail how the causal knowledge helps the Visual language Model or reasoning model to improve its efficiency in question answering task related to music videos.

\section{Related Work}
\subsection{Multimodal and Audio-Visual Understanding}
KARMA-MV operates across three modalities, hence is at the intersection of multimodal representation learning and audio-visual understanding. Sun et al. in~\cite{Sun_Sarma_Sethares_Liang_2020} proposed Interaction Canonical Correlation Network (ICCN), for multimodal sentiment analysis and emotion recognition by jointly modelling text, audio, and video features. 
\subsection{Video Question Answering and Causal Reasoning}

A broad range of video QA datasets has been developed for video understanding~\cite{survey}, spanning story comprehension~\cite{movieqa}, TV show understanding~\cite{tvqa}, and general web videos~\cite{activitynet}. Among these, the datasets most relevant to our work are those that target causal or temporal reasoning rather than simple factual retrieval.

NExT-QA~\cite{nextqa} provides MCQs for causal and temporal action reasoning in video, establishing an important precedent for structuring causal inference as a multiple-choice task. Most directly related is Causal-VidQA~\cite{li2022representation}, which introduced a large-scale benchmark of 107,600 QA pairs designed to advance video understanding into causal and commonsense reasoning. The dataset defines four question types---description, explanation, prediction, and counterfactual---which we adopt in KARMA-MV. However, Causal-VidQA focuses entirely on visual content, whereas KARMA-MV is specifically concerned with how visual changes causally drive changes in music and audio. The Audio-Visual Question Answering dataset (AVQA)~\cite{li2022dynamic} addresses questions about sounds, visual objects such as instruments, and their associations in video, making it the closest prior work in the audio-visual QA space to our own. Our benchmark extends this direction by explicitly targeting the causal direction from video to music, introducing counterfactual and prediction question types absent from AVQA.

\section{Methodology}\label{sec:page_size}
The work is broadly divided into two parts: \textbf{I. Dataset Creation and testing on baseline VLMs. II. Building Knowledge Graph VLM architecture.}
The dataset creation pipeline comprises four steps: 1) Feature extraction, 2) Generation of causal reasoning dataset 3) Generation of MCQ dataset 4) Validating the dataset. Our approach is very simple. Unlike many other datasets, KARMA-MV does not rely on human annotators for either creation or validation. Instead, we have leverage an LLM for automated dataset creation and VLMs for baseline evaluation, which ensures scalability without sacrificing semantic richness.
After validation, we proceed to build a Causal Knowledge Graph augmented VLM architecture to answer the MCQ questions in the KARMA-MV dataset.
For the dataset construction, we assemble a corpus of music videos available on YouTube under Creative Commons licenses, totalling of 2,682 music-videos. These are segmented into clips using PySceneDetect~\cite{Castellano_PySceneDetect} for scene transition analysis in the following steps.

\subsection{Feature Extraction}\label{subsec:body}
Two broad categories of features are extracted: 1)\textbf{ Music and Audio Features} and 2) \textbf{Visual Features.} 
After segmenting music videos, we filter the resulting clips for quality and relevance, yielding a set of 22,011 music-video scenes. For each scene, we extract a comprehensive set of features including loudness, tempo, key, chord, mood, genre, instruments used, and voice detection using multiple specialized libraries. Similarly, visual features such as object count and visual intensity are extracted using separate libraries.

\textbf{Music and Audio Features}. For loudness extraction we use Librosa ~\cite{librosa_spicy}, a standard python library for analyzing audio and music signals. For mood extraction we use Music2Emo~\cite{kang2025unified}. For tempo, genre, chord, instrument detection and voice detection, we follow~\cite{mirflex2024}. For key,genre and instrument extraction we rely on the Essentia Library~\cite{alonso2020essentia}.


\textbf{Visual Features}. To quantify the visual characteristics of the video stimuli, we extract four key metrics: Motion, Brightness, Contrast, and Saturation, by sampling every fifth frame to balance granularity with computational efficiency. Motion is calculated via frame-differencing after applying a Gaussian blur to reduce noise, while the remaining aesthetic features are derived from the HSV (Hue, Saturation, Value) color space. These metrics are then aggregated into a Total Visual Intensity (TVI) score using a weighted linear combination: TVI = 0.5M + 0.3C + 0.2S. This composite metric represents the overall dynamic energy of each scene, providing a standardized basis for subsequent statistical analysis.
Beyond visual characteristics, we use YOLOv8-Medium (YOLOv8m) pre-trained model~\cite{yolov8_ultralytics} for number of types of objects. For each video segment, inference is performed on the middle frame to obtain a representative snapshot of the scene's content.
After feature extraction, we merge all the audio/music features and video features into per-video feature files, where each row corresponds to one scene. These raw features serve as the input for dataset preparation and causal analysis.

\subsection{Generating Causal Reasoning Dataset}
We leverage Qwen2.5-7B-Instruct~\cite{qwen2.5,qwen2_tech} to construct a causal reasoning dataset over scene transitions in the music video. Our causal reasoning is grounded in how the visual delta i.e. change in visual features affect the music or audio spectrum. Causal reasoning statements are generated in natural language by feeding the raw extracted features of each pair of consecutive scenes directly to the LLM.. Each entry in the causal reasoning dataset captures three components required for downstream MCQ generation:
\begin{itemize}[nosep, leftmargin=*] 
\item \textbf{Audio/Music Delta}: the change in the audio and music properties occurring across the transition between two consecutive scenes. 
\item \textbf{Visual Delta}: change in visual features occurring across the transition between two consecutive scenes.
\item \textbf{Causal Reasoning}: the central component our analysis, which focuses on the relation between visual delta and the resulting musical change. This gives us a very natural-language explanation about how the change in the visual cues affects the music. 
\end{itemize}
Counterfactual questions in the MCQ dataset are subsequently framed around hypothetical scenarios derived from these causal links, with the LLM constructing hypothetical premise from the causal reasoning provided as input.

\subsection{Generating the KARMA-MV MCQ Dataset}
The causal reasoning generated in the previous step serves as a reference data for constructing the MCQ dataset. Questions are formulated to be concise, understandable and interpretable to human actors. The qwen-2.5-7B-Instruct LLM is prompted to generate questions grounded in reasoning statements from the previous step. Following Li et al.~\cite{li2022representation}, each scene transition yields 3 MCQs, each having 4 options, of which only one is correct. The question types are:
\begin{itemize} [nosep, leftmargin=*] 
    \item \textbf{Evidence Reasoning}: questions that probe the deeper causal relationship between changes in the visual domain and their effects on music, requiring the model to identify the correct causal mechanism.
    \item \textbf{Prediction}: questions that ask how the music will change across a scene transition given the observed change in visual content.
    \item \textbf{Counterfactual}: questions that posit a hypothetical scenario in the context of the scenes and ask what musical outcome would follow, testing the model's ability to reason about alternative causal chains.
\end{itemize}

The resulting KARMA-MV dataset contains 12,579 MCQ questions with music video for each category, resulting in a total of 37,737 MCQs. 


\section{Causal model}

\subsection{Causal Knowledge Graph Construction}
\begin{figure}
    \centering
    \includegraphics[width=\columnwidth]{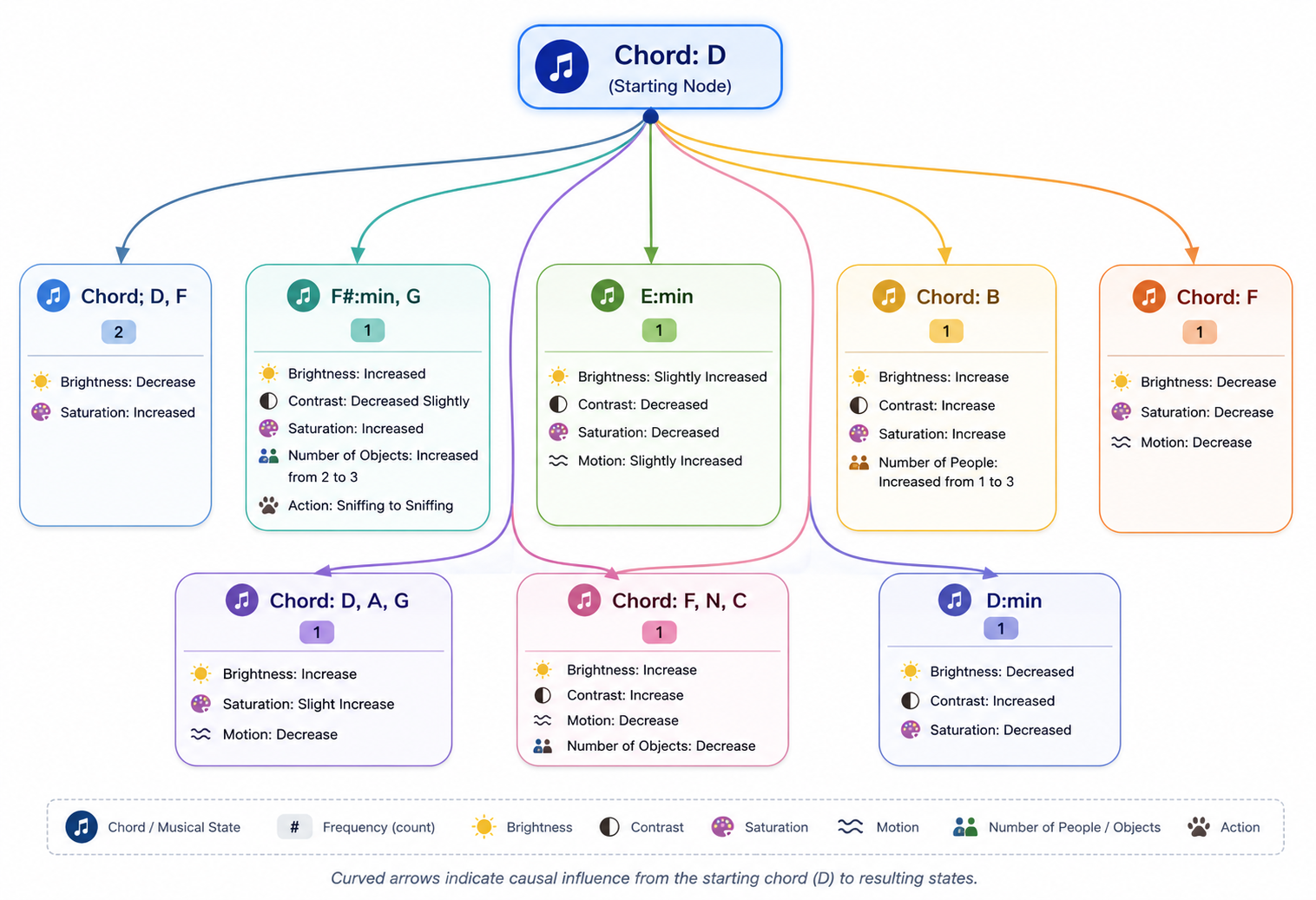}
\vspace{-6pt}
    \caption{\footnotesize Excerpt from the causal knowledge graph. }
    \label{fig:causal}
\vspace{-12pt}
\end{figure}

To model causality explicitly and improve downstream model performance, we augment our VLM baselines with Causal Knowledge information encoded as structured graph representations. A Causal Knowledge Graph (CKG) is a natural representation for this purpose, as it enables efficient search, retrieval, and reuse of causal facts across experiments. Following Xu et al.~\cite{Xu19052023},
we construct our causal knowledge based model from the causal reasoning dataset. In this graph, the nodes represent Music feature states and directed edges represent changes in visual features that cause transitions in music feature states. We prompt the same base LLM, Qwen2.5-7B-Instruct, to generate causal triplets from the causal reasoning dataset: for each transition pair, a triplet is produced comprising a \textit{Source} (Change in Visual cues), \textit{Relation} (brief reasoning on how change in video affects music/audio), and \textit{Target} (change in Music/Audio features). 
These causal triplets serves as ground truth from which final Causal Knowledge Graph is constructed using NetworkX~\cite{NetX}. An extract of the resulting graph is provided in Figure \ref{fig:causal}.

The constructed causal graph comprises 9,258 nodes and 13,101 edges. While the low graph density (0.00015) and modest average degree (1.42) are consistent with the principle of causal parsimony \cite{pearl2010causal}, these metrics specifically define a music state transition network where edges reflect directional, temporally ordered shifts in genre, mood, and instrumentation[cite: 259, 260]. Our topological analysis reveals that nearly half of these music states (48.57\%) function as causal initiators or root nodes with zero in-degree, suggesting they act as spontaneous origins of musical sequences rather than derivatives of prior states.

Within this structure, ``Genre: Pop'' and ``Instruments: Changed'' emerge as the primary catalysts for musical evolution, initiating the highest number of downstream transitions (129 and 119, respectively). Furthermore, the decomposition of the graph into 1,357 distinct causal communities suggests the existence of self-contained musical sub-systems with limited cross-community flow. The largest of these clusters encompasses 3,380 nodes, or 36.5\% of the total graph, indicating a dominant interconnected structure likely corresponding to mainstream popular music styles, which exists alongside a long tail of smaller, niche stylistic communities.

\subsection{VLM grounding with CKG}
Having constructed the Causal Knowledged Graph, we integrate it with Baseline VLMs to evaluate whether access to structured causal knowledge improves question answering performance. The core idea is to retrieve the most relevant causal facts from the graph for each MCQ at inference time and inject them into the model's prompt as additional context. To enable efficient retrieval, the raw graph is serialized into column-oriented Parquet files and decomposed into distinct Entity and Relationship dataframes using pandas and NetworkX~\cite{NetX}. 
We adopt the 1-hop retrieval strategy operating over a NetworkX directed graph. We first extract semantic keywords from the question text and perform a topological search to identify matching anchor nodes within the graph. From each anchor node, we traverse only outgoing edges, which by construction ensures that retrieval captures forward-looking causal effects rather than backward-looking historical causes. We score each retrieved relationship using a bipartite heuristic scoring function: the base weight is derived from the anchor node's degree centrality; naturally prioritizing highly connected, authoritative musical concepts; while a 1.5x transition boost is applied if the edge description contains kinetic terminology (e.g., increase, shift). Finally, we rank the candidate facts by this composite score and truncate to a top-$k$ list, where $k$ can be tuned per model to manage the information bottleneck. In our experiment with set $k$ to 25 for VLMs and 3 for thinking LLMs. We append the resultant causal context to the VLM prompts when answering MCQs in KARMA-MV dataset.

\subsection{CKG Integration with LLM}
For integrating our CKG facts into a thinking and reasoning Language Model, we use structured Zero-shot Chain of Thought (CoT) prompting technique. First we use NetworkX to develop a directed graph, from which causal facts are retrieved using keywords extracted from each MCQ for the given transition pair. A well-defined feature set is then assembled and provided to the LLM as a structured prompt. Specifically, the architecture retrieves relevant information from the graph and feeds them to the LLM alongside the ground truth feature-evidences for the transition pairs, and the LLM is then asked to answer the questions based on these information. 

\begin{table*}[t]
\centering
\footnotesize
\setlength{\tabcolsep}{2pt} 
\renewcommand{\arraystretch}{1} 
\begin{tabular}{llcccc}
\toprule
\textbf{Model Architecture} & \textbf{Type} & \textbf{Evidence Reasoning} & \textbf{Prediction} & \textbf{Counterfactual} & \textbf{Overall} \\
\midrule
\multicolumn{6}{l}{\textit{Without Knowledge Graph Grounding}} \\
\hspace{3mm} Qwen-2.5-Omni-7B & VLM & 65.73 & 76.57 & 56.82 & 66.37 \\
\hspace{3mm} MiniCPM-o-4.5 & VLM & 74.06 & 71.11 & 69.39 & 71.51 \\
\hspace{3mm} Gemma-4-31B Instruct & LLM & 78.20 & 73.37 & 74.11 & 75.22 \\
\midrule
\multicolumn{6}{l}{\textit{With Causal Knowledge Graph (CKG) Grounding}} \\
\hspace{3mm} Qwen-2.5-Omni-7B + CKG & VLM & 73.35 & 78.90 & 68.98 & 73.74 \\
\hspace{3mm} MiniCPM-o-4.5 + CKG & VLM & 77.69 & 71.67 & 72.90 & 74.09 \\
\hspace{3mm} Gemma-4-31B Instruct + CKG & LLM & 79.17 & 76.68 & 72.29 & 76.05 \\
\bottomrule
\end{tabular}
\caption{ \small Accuracy (\%) of the Multiple Choice Question (MCQ) task highlighting the performance gains achieved by integrating Causal Knowledge Graph (CKG) facts into VLM and LLM architectures. Random chance baseline is 25\%. We note that VLM models are provided the raw music video, whereas the LLM model is provided with the extracted features.}
\label{tab2}
\end{table*}
\begin{figure*}[htbp]
    \centering
    \begin{minipage}[t]{0.44\textwidth}
        \centering
        \begin{subfigure}{\linewidth}
            \centering
            \includegraphics[width=\linewidth]{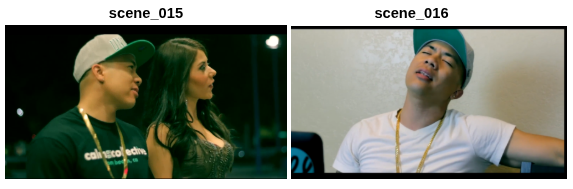}
            \vspace{-6pt}
            \caption{\footnotesize Transition scene pair (input)}
            \label{fig:sub_trans1}
        \end{subfigure}
        \vspace{1ex} 
        \begin{subfigure}{\linewidth}
            \centering
            \includegraphics[width=\linewidth]{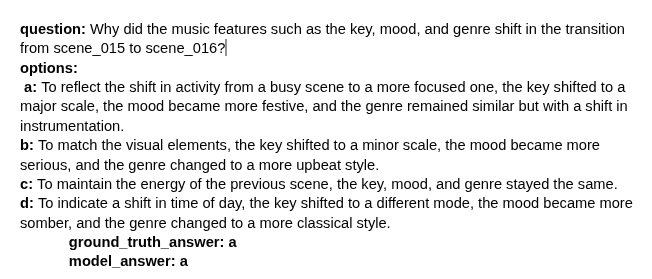}
            \caption{ \footnotesize Evidence reasoning question and CKG-grounded response}
            \label{fig:sub_evid}
        \end{subfigure}
    \end{minipage}
    \hfill
    \begin{minipage}[t]{0.44\textwidth}
        \centering
        \begin{subfigure}{\linewidth}
            \centering
            \includegraphics[width=\linewidth]{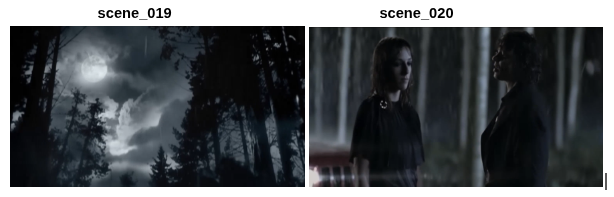}
            \caption{\footnotesize Transition scene pair (input)}
            \label{fig:sub_trans2}
        \end{subfigure}
        \vspace{1ex} 
        \begin{subfigure}{\linewidth}
            \centering
            \includegraphics[width=\linewidth]{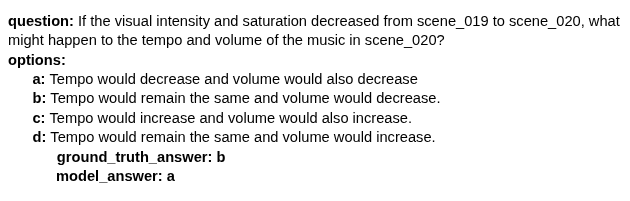}
            \caption{ \footnotesize Prediction question and CKG-grounded response}
            \label{fig:sub_pred}
        \end{subfigure}
    \end{minipage}
    \vspace{-6pt}
    \caption{\small Qualitative examples of responses generated by Qwen-2.5-Omni-7B integrated with the CKG on evidence reasoning and prediction question types.}
    \label{fig:main_figure}
    \vspace{-10pt}
\end{figure*}

\section{Experimental setup}

We evaluate the baseline performance of current open VLMs and thinking LLM's on KARMA-MV MCQs, and subsequently ground them with the developed Causal Knowledge Graph. 
We include two state-of-the-art VLM models in our experiment: Qwen-2.5-Omni-7B~\cite{xu2025qwen25omnitechnicalreport} and MiniCPM-o 4.5~\cite{yao2024minicpm}. During evaluation, the raw music-video transition scene pairs are directly fed to the VLM and subsequently the questions related to this particular transition pair are asked in the prompt. For prediction-type questions, where the model is asked to predict the music in the second scene, we mute the audio/music track of the second scene using ffmpeg to prevent information leak. 



Finally, we use the Gemma-4-31B Instruct LLM~\cite{gemma_4_31b_it} as a thinking model on our dataset. As this model cannot consume the raw music video files, we instead provide the extracted music, audio, and visual features for each transition pair as structured text input. First we establish a baseline and then evaluate its performance when augmented with CKG facts following the retrieval and injection procedure described in the methodology.




\section{Results and Discussion}

\subsection{Improving MCQ answering with CKG}

\subsubsection{Baseline results}
Examining the baseline models provides a picture of how the VLMs behave on our dataset without any external causal grounding. As shown in Table \ref{tab2}, Qwen 2.5 Omni achieves an overall accuracy of 66.37\%, the lowest among the three baselines. Counterfactual questions prove to be its most significant weakness, with an accuracy of only 56.82\%; barely above random chance for a 4-option MCQ; while prediction-type questions are comparatively more tractable (76.57\%). This pattern is consistent with the nature of the task: prediction questions require forward temporal inference from observed visual context, whereas counterfactual questions demand reasoning over hypothetical scenarios with no direct perceptual grounding in the video, a capacity that appears limited in a 7B VLM without external causal context. 

MiniCPM-o-4.5 performs substantially better overall (71.51\%), with a more balanced accuracy profile across categories (Evidence: 74.06\%, Prediction: 71.11\%, Counterfactual: 69.39\%). Notably, while Qwen shows a pronounced weakness in counterfactual reasoning, MiniCPM's counterfactual performance (69.39\%) is within 5 points of its evidence reasoning score, suggesting that its instruction-tuning and thinking mode contribute to more robust causal generalization.

The best-performing baseline is Gemma-4-31B Instruct (75.22\% overall), with relatively strong counterfactual performance (74.11\%). It is important to note, however, that Gemma operates on structured feature representations rather than raw video, which simplifies the inference problem and reduces the perceptual complexity of the task. This input modality difference makes direct comparison with the VLMs somewhat asymmetric, though it does not diminish the validity of the CKG integration analysis conducted separately for each model.

\subsubsection{Causal Knowledge Graph Grounded Architecture}
Integrating CKG facts into the model prompts yields consistent accuracy improvements, with the magnitude of improvement varying substantially across models and question types as shown in Table \ref{tab2}.

Qwen-2.5-Omni-7B, benefits most from CKG augmentation, gaining 7.37 percentage points overall (from 66.37\% to 73.74\%). The category-level breakdown reveals a particularly striking result: the largest absolute gain occurs on counterfactual questions, where accuracy improves by 12.16 points (from 56.82\% to 68.98\%). This is precisely the question type where Qwen's baseline was weakest, suggesting that the model lacked the internalized causal knowledge necessary to reason about hypothetical audio-visual scenarios, and that the CKG directly compensates for this deficit by supplying the relevant causal links at inference time. Evidence reasoning also improves substantially ($+$7.62 points), while prediction questions show a more modest gain ($+$2.33 points), consistent with the baseline already being strong in that category.

MiniCPM-o-4.5, augmented with the CKG, improves by 2.58 points overall (71.51\% to 74.09\%). Unlike Qwen, the gains are distributed more uniformly across question types: evidence reasoning improves by 3.63 points, counterfactual by 3.51 points, and prediction by a marginal 0.56 points. This uniform improvement profile suggests that for a model with stronger baseline causal reasoning capacity, the CKG provides consistent contextual grounding rather than targeted remediation of a specific reasoning weakness. The smaller absolute gain compared to Qwen is expected given MiniCPM's stronger starting point.

For Gemma-4-31B Instruct, the CKG yields the smallest overall improvement (+0.83 points, from 75.22\% to 76.05\%), with gains concentrated in evidence reasoning (+0.97) and prediction (+3.31) categories. Notably, counterfactual accuracy \emph{decreases} by 1.82 points (from 74.11\% to 72.29\%) when the CKG is integrated. We observe that when the thinking LLM is given too many facts, it tends to hallucinate or over-rely on the provided context, which disrupts the logical consistency required for counterfactual inference. This suggests that for large models with rich internalized causal knowledge, injecting excessive graph-derived constraints can restrict rather than support their capacity for open-ended hypothetical reasoning.

\subsubsection{Model Capacity and CKG Benefit: An Inverse Scaling Observation}

A consistent pattern emerges across all three models: the benefit from CKG augmentation is inversely related to model capacity. The 7B Qwen model gains 7.37 percentage points overall, the 9B MiniCPM gains 2.58 points, and the 31B Gemma gains only 0.83 points. This trend is consistent with the hypothesis that larger models internalize more causal world knowledge during pre-training, reducing the marginal contribution of explicit external knowledge at inference time. Conversely, smaller models operating on tasks that demand causal reasoning beyond their parametric capacity stand to gain the most from structured knowledge injection.

This observation has a practical implication for benchmark design and system deployment: CKG augmentation is a particularly effective compensatory mechanism for resource-constrained inference settings, where deploying a smaller model augmented with a lightweight knowledge graph may approach or match the performance of a larger model without graph grounding. In our experiments, Qwen-2.5-Omni-7B with CKG (73.74\%) substantially narrows the gap to the unaugmented MiniCPM-o-4.5 (71.51\%) and approaches MiniCPM with CKG (74.09\%), despite having fewer than half the parameters.

\begin{figure}[h!]
    \centering
    \includegraphics[width=0.8\columnwidth]{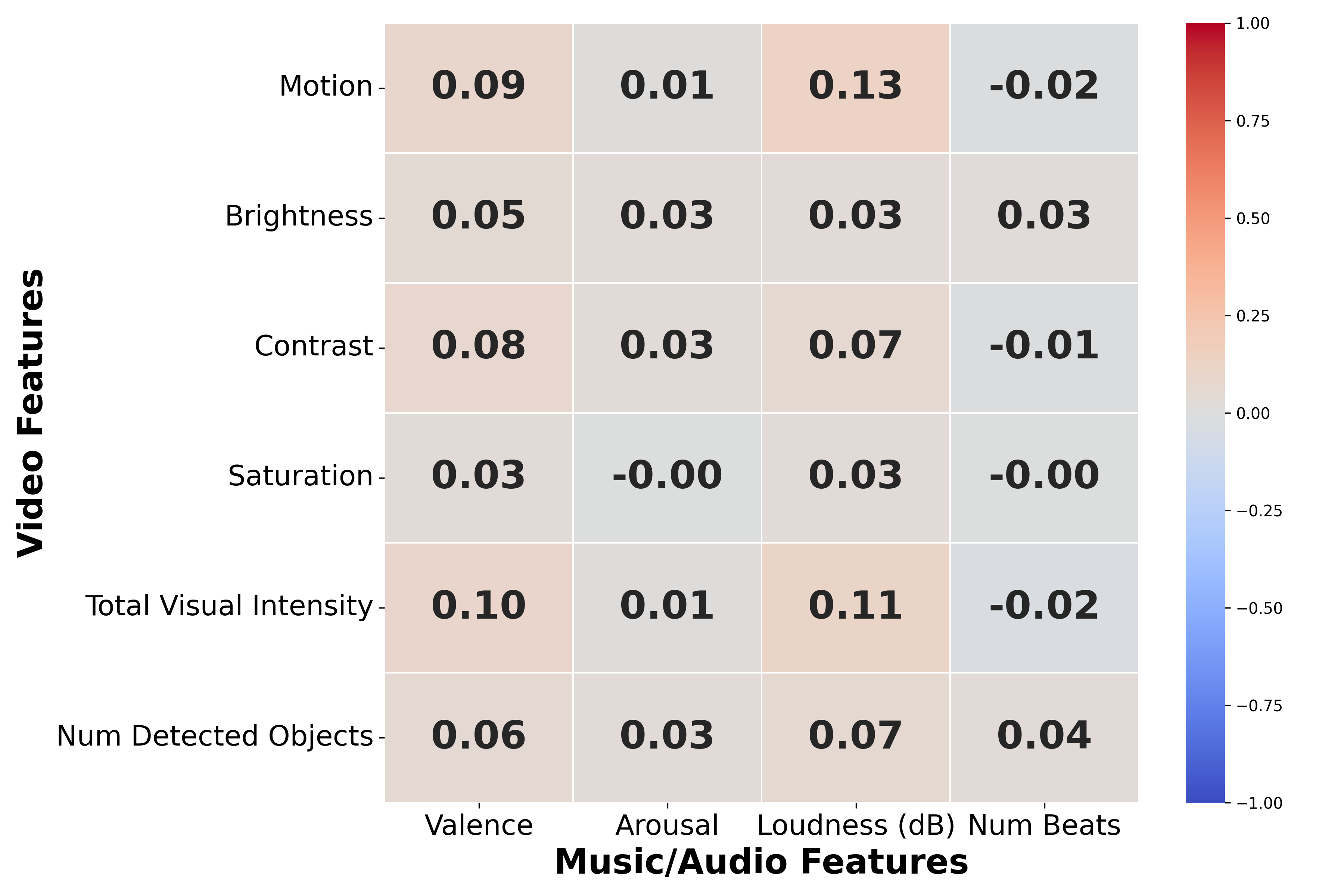}
    \vspace{-6pt}
    \caption{\small Spearman rank correlation heatmap illustrating cross-modal dependencies between  video features (y-axis) and audio/music features (x-axis)}
    \label{fig:heatmap}
    \vspace{-12pt}
\end{figure}
\subsection{Cross-modal Dependencies in Dataset}
To evaluate audio-visual dependencies, we compute Spearman rank correlations across all 22,011 scenes (Figure~\ref{fig:heatmap}). The positive correlations between Total Visual Intensity (TVI) and loudness, and between motion and arousal, provide quantitative validation that KARMA-MV is grounded in genuine cross-modal dependencies rather than incidental co-occurrence or dataset artifacts. Furthermore, the correlation between visual saturation and valence aligns with prior work on color-affect mapping, suggesting that even low-level visual features contribute to audio-visual coherence. Together, these patterns establish a quantitative baseline for benchmarking future causal frameworks in the music-video domain.




\subsection{Discussion and Future Work}


Causal reasoning about music-video relationships is inherently difficult even for trained human annotators: correctly attributing a change in musical key or rhythmic density to a specific visual shift requires simultaneous expertise in music theory and visual analysis, making human annotation unreliable. This is precisely why KARMA-MV adopts an automated pipeline grounded in high-accuracy specialized feature extractors and a well-grounded state-of-the-art LLM, trading the noise of lay annotation for consistent, reproducible feature-grounded generation.


Future work could address this through adaptive retrieval strategies that modulate the amount of injected causal context based on model size or confidence, reducing the anchoring effect for larger models. Additional directions include richer causal interventions beyond scene transitions, broader genre coverage, and more expressive graph representations that better capture the graded nature of audio-visual causal influence.


\section{Conclusion}
We introduced KARMA-MV\footnote{\url{https://huggingface.co/datasets/amaai-lab/Karma-MV}}, a large-scale benchmark for causal question answering in music videos, targeting how changes in visual content drive changes in music and audio. The dataset is constructed entirely through automated, feature-grounded generation, reflecting the genuine difficulty of the task for human annotators. Experiments across three state-of-the-art models demonstrate that explicit causal knowledge graph grounding yields consistent gains---particularly for smaller models, where the CKG most directly compensates for limited internalized causal knowledge. The inverse scaling pattern observed, where larger models benefit less from external causal context, suggests that structured knowledge injection is most valuable in resource-constrained deployment settings. The dataset, causal knowledge graph as well as the VLM/LLM setup are available open-source\footnote{\url{https://github.com/AMAAI-Lab/Karma-MV}}.

KARMA-MV establishes a scalable testbed and strong baseline for causal audio-visual reasoning, with clear directions for future work in adaptive knowledge retrieval, richer causal interventions, and broader genre coverage.

\section{Acknowledgments}
This work has received funding from grant no. SUTD SKI 2021\_04\_06 and from MOE grant no. MOE-T2EP20124-0014

\section{AI Usage Statement}
We acknowledge the use of Gemini and Claude for grammar improvements.  

\bibliography{ISMIRtemplate}

%
%
%
%

\end{document}